# Robopinion: Opinion Mining Framework Inspired by Autonomous Robot Navigation

M. A. El-Dosuky[1], M. Z. Rashad[1], T. T. Hamza[1], A.H. EL-Bassiouny[2]

[1] Dep. of Computer Sciences, Faculty of Computers and Info. Mansoura University, Egypt,
[2] Dep. of Mathematics, Faculty of Sciences, Mansoura University, Egypt
[1]{mouh_sal_010, magdi_12003,Taher_Hamza}@mans.edu.eg, [2]el_bassiouny@mans.edu.eg

**Abstract.** Data association methods are used by autonomous robots to find matches between the current landmarks and the new set of observed features. We seek a framework for opinion mining to benefit from advancements in autonomous robot navigation in both research and development.

## 1 Introduction

Opinion mining (OM) is a new field of data mining. Its underpinnings are information retrieval and appraisal theory which is a recent theory in computational linguistics [1]. OM is concerned with the opinion that can be induced from documents. OM is divided into three major tasks: development of linguistic resources, sentiment classification, and opinion summarization, as shown in figure 1 [2].

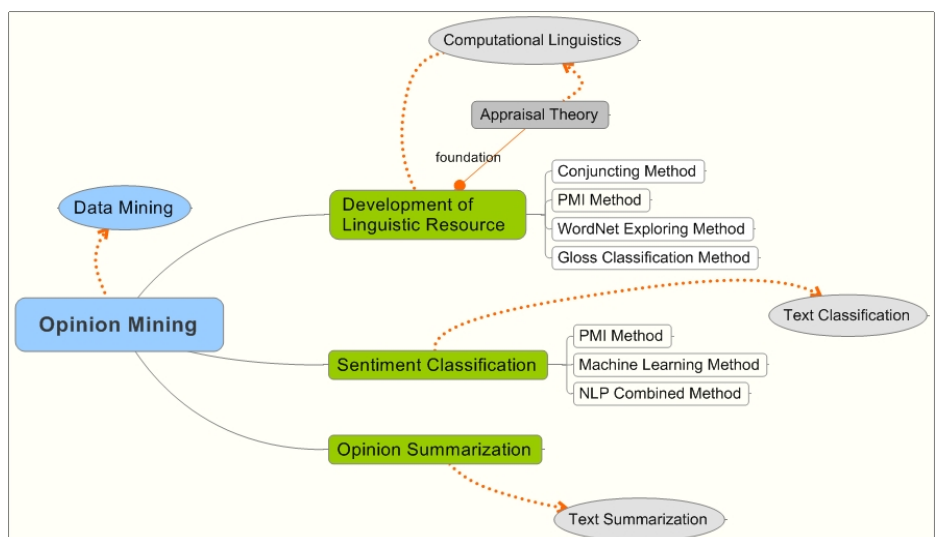

**Fig. 1.** Tasks for opinion mining and its relationship with related area

We seek a framework for OM that benefits from advancements in autonomous robot navigation in both research and development. For a mobile robot to be autonomous, it must solve the simultaneous localization and mapping (SLAM) problem, i.e. to be able to progressively build a consistent map of an unknown environment while simultaneously determining its location within this dynamically changing map [3]. The solution of SLAM problem, especially the data association task has been one of the notable successes of the robotics community over the past decade, as surveyed in [3]. Data association is very crucial task to find matches between the current landmarks and the new set of observed features [4].

The utilization of data association methods for OM is not a new idea. Pointwise mutual information (PMI) has been used for determining Semantic Orientation (SO) of a term with high accuracy within very large corpora [5]. PMI calculates the co-occurrence probability of each term pair $x, y$.

$$PMI(x, y) = \log_2 N \frac{p(x, y)}{p(x)p(y)} \tag{1}$$

SO of a term can be calculated by considering some paradigmatic terms.

$$SO(t) = \sum_{t_i \in Pos} PMI(t, t_i) - \sum_{t_i \in Neg} PMI(t, t_i) \tag{2}$$

where $t$ is the target term and $t_i$ is a paradigmatic term. $SO(t)$ can be either positive or negative, and the magnitude is the strength of the orientation.

The rest of the paper is outlined as follows. Section 2 reviews the tasks for opinion mining. Section 3 introduces Robopinion – a framework for OM that benefits from advancements in the community of autonomous robot navigation. Section 4 investigates the proposed framework and concludes the paper.

## 2    Background

### 2.1    Opinion Mining Tasks

First, we need defining some sentiment-related properties of terms such as term orientation (Positive/Negative), term subjectivity (Subjective/Objective), and the strength of term attitude. Developing linguistic resources can be achieved in many ways. The major methods are: PMI [5], conjunction method [6], WordNet exploring method [7], and gloss classification method [8]. SentiWordNet is developed as publicly available linguistic resource [9].

To determine polarity, sentiment classification is applied to classify the sentiment-orientated documents into positive or negative. Mean semantic orientation of extracted phrases in a given document can determine the category the document belongs to [10]. An alternate approach is to apply standard machine learning classification techniques for classifying the sentiment [11]. In this approach, a document is converted into vector of high dimensionality before applying classifiers, such as: SVM, maximum entropy, Naïve Bayes, and ADTree.

Finally, Results are summarized and visualized in many forms. Some systems use thumbs up/down such as *ReviewSeer*[12], and some use star rating such as *RedOpal*[13]. Other systems use linguistic resources with sentiment lexicons, such as *Opinion Observer*[14], and *WebFountain*[15].

### 3.2 Data Association for Robot Navigation

The general architecture of the control unit of most autonomous robots is shown in the next figure [16]. In stochastic mapping [17], data association is simply addressed using a classical technique in tracking problems called the nearest neighbor (NN) [18]. The development of the robust method of Joint Compatibility Branch-and-Bound data association (JCBB) is encouraged by the fact that the simple Nearest Neighbor (NN) algorithm for data association is very sensitive to the sensor error, increasing the probability of matching unrelated map features [4].

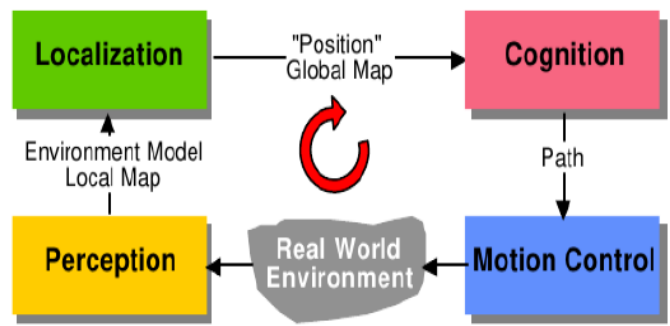

**Fig.2.** Control Architecture of autonomous robots

José Neira builds a MATLAB™ simulation that demonstrates data association methods for a mobile robot equipped with a point detector [19]. The robot is to move in an environment with red point features at each side of a square trajectory, where red points and trajectory are ground truth.

## 3 Proposed Framework

### 3.1 Robopinion

The main procedure is shown in the following figure.

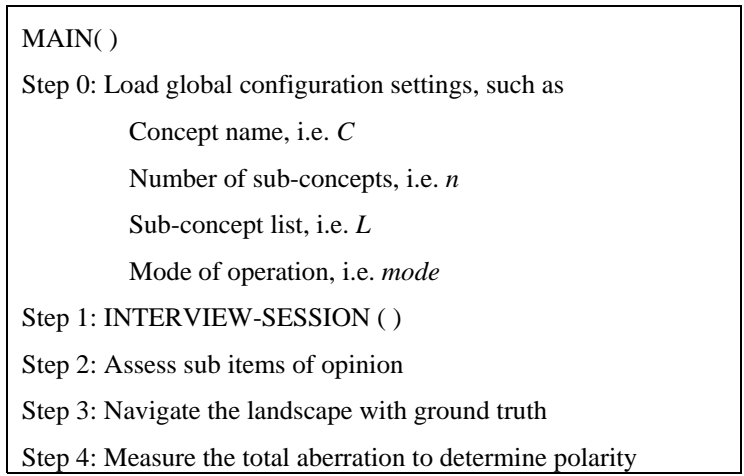

Fig.3 main procedure

A concept name *C*, such as a new product to express an opinion about, has a number of sub-concepts *L* such as price, weight, etc. The concept provider is called imposer, as it is his/her responsibility to provide the set of features' evaluation criteria.

Step 1 of the main procedure introduces interview sessions to get opinion documents. Interviews are one of simplest knowledge acquisition methods. A structured interview forces an organization of communications with asking planned questions, thus reducing problems inherent in interviews and helps preventing the distortion caused by vagueness and subjectivity [20]. INTERVIEW-SESSION is shown in next figure.

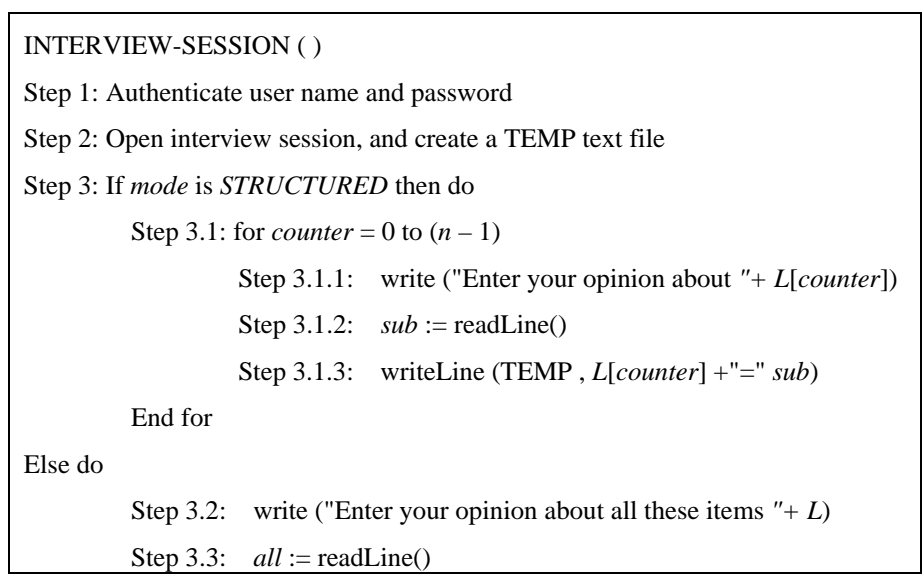

```
            Step 3.4: for counter = 0 to (n – 1)
                    Step 3.4.1:   sub := concordance(L[counter], all)
                    Step 3.4.2:   writeLine (TEMP , L[counter] +"=" sub)
            End for
    End if
    Step 4: close the interview session, and save the TEMP text file
```

Fig.4 Interview session procedure

Text concordance, mentioned in step 3.4.1, is one of the most commonly used corpus analysis tools, and certainly the oldest to obtain a table in which words which occur in a text are paired with citations of the text passages in which they occur. The art of concordancing hits the highest point in computational corpus linguistics [21].

Step 2 of the main procedure introduces assessment method. We adopt Likert scale [22] as it is a common psychometric scale widely used to scaling responses in survey research that employs questionnaires [23]. We represent the assessment as plum representation shown in the next figure.

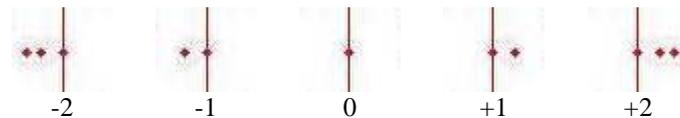

$-2 \quad -1 \quad 0 \quad +1 \quad +2$

Fig 5 plum representation of polarity

The assessment itself touches some linguistic topics. First, copulas play an important role in the categorization of the lexicon [24]. If we encounter a phrase in the form "$X$ is $Y$" where $X \in L$, then we can say with a great confidence that the assessment is the evaluation of $Y$ using appraisal theory [1]. Another linguistic concern is inflectional morphology, especially declension of nouns, adjectives and pronouns [25]. If X' is an inflection of X then it can be replaced by X.

Step 3 of the main procedure introduces robot navigation in a landscape. This may utilize JCBB data association [19]. Converting textual data to numerical coordinates can be done using a technique with high speed up [26]. The construction of the landscape may obey the following assumptions for fair judgement:
- The imposer should state an even number of sub-concepts, here we have 10.
- Odd numbered sub-concepts represent positive features
- Even numbered sub-concepts represent negative features
- The total sum of all features is zero.
- Two dummy nonnegotiable sub-concepts are added: the zero and pre-zero.
- The zero sub-concept denotes the existence of a concept in hand.
- The pre-zero sub-concept: "Cogito ergo sum"; or "I think, therefore I exit".

All these considerations are shown in the landscape in the next figure. Ground truth points are usually imposed, but can be obtained from the so-for established facts from previous interview sessions.

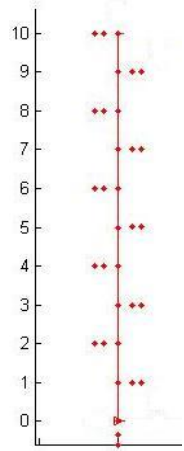

**Fig. 6** ground truth

### 3.3 Results

Figure 7 shows the trail and the depiction of the normalized aberration in x axis for the ten features by exploiting JCBB data association. Aberration in x axis is made to be normalized that the maximum negative aberration subtracted from the maximum positive aberration gives one.

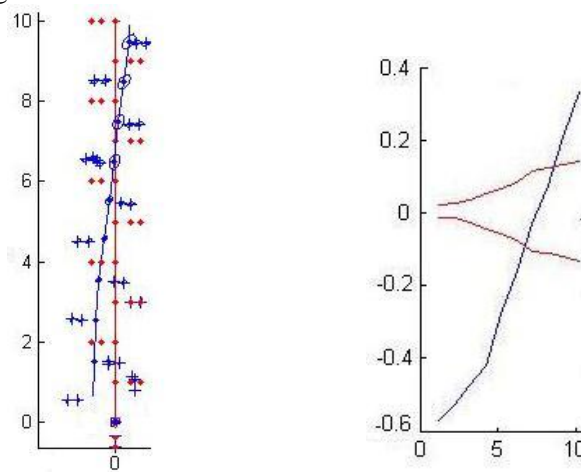

Fig. 7 Trail and normalized aberration in x axis using JCBB data association

With notice three slopes: the positive aberration slope with length 0.31, the negative aberration slope with length 0.6, and the in-between neutral slope with length 0.09. For that session we can say that the opinion polarity is negative as -0.29 = 0.31 – 0.6.

## 4   Discussion

The main goal of this paper is to extract, classify, and understand opinions. We developed a framework for OM that benefits from advancements in autonomous robot navigation in both research and development.   Many researchers consider the topic of SLAM to be close to being solved. Data association task has been one of the notable successes of the robotics community over the past decade. Data association is very crucial task to find matches between the current landmarks and the new set of observed features.

The future work may have many directions. One possible direction is studying the possibility and the logic behind getting opinion about the existence of something. The zero sub-concept denotes the philosophical consideration that ontology precedes epistemology. As we start with the zero assumption that the thing exist, getting a negative polarity may be a paradox. Consider the political debate corpus [27], the "God" folder. Are we studying the opinion about existing things? at least in the minds of the imposer and the opinion holders.

Another possible direction is studying *"the benefit of the doubt"* due to vagueness. The benefit of the doubt is common term in judiciary jargon that is to believe something good, rather than something bad, when you have the possibility of doing either. Again, let us consider the political debate corpus. Within the "God" folder, we have the following two posts:

    post_209
    #stance=stance2 #originalStanceText=No #originalTopic=is-there-a-god
    Exactly

    post_940
    #stance=stance2 #originalStanceText=against #originalTopic=God_Exists
     Forty Two

The first is clear, while the other is vague, referring to The Hitchhiker's Guide to the Galaxy, a science fiction comedy series created by Douglas Adams.

## References

1. Martin, J.R., White Peter R. R, The Language of Evaluation: Appraisal in English. 2005: Palgrave Macmillan
2. Dongjoo Lee et al., Opinion Mining of Customer Feedback Data on the Web, 2008, Proc. of The $2^{nd}$ Intr'l Conf. on Ubiquitous Info. Management and Communication, pp 247-252
3. Hugh Durrant-Whyte, Tim Bailey, "Simultaneous Localization and. Mapping: Part I", IEEE Robotics & Automation Magazine, June. 2006.
4. Neira, J. Tardos; Tardos, J.D., "Data association in stochastic mapping using the joint compatibility test", IEEE Transactions on Robotics and Automation, vol.17, issue 6, pp 890–897, 2001
5. Peter D. Turney, M.L.L., Measuring Praise and Criticism: Inference of Semantic Orientation from Association. ACM Transactions on Information Systems, 2003. 21(4): p. 315-346.
6. Vasileios Hatzivassiloglou, K.M. Predicting the Semantic Orientation of Adjectives. in EACL. 1997: ACL.

7. Minqing Hu, B.L. *Mining and Summarizing Customer Reviews*. in *Tenth ACM SIGKDD International Conference on Knowledge Discovery and Data Mining*. 2004. Seattle, Washington, USA: ACM.
8. Andrea Esuli, F.S. Determining Term Subjectivity and Term Orientation for Opinion Mining. in EACL. 2006. Trento, Italy: The Association for Computer Linguistics.
9. Andrea Esuli, F.S. SentiWordNet: A Publicly Available Lexical Resource for Opinion Mining. in LREC-06. 2006. Genova, IT.
10. Turney, P.D. Thumbs Up or Thumbs Down? Semantic Orientation Applied to Unsupervised Classification of Reviews. in ACL (40th Annual Meeting of the Association for Computational Linguistics). 2002. Philadelphia, Pennsylvania, USA: ACL.
11. Bo Pang, Lillian Lee, Shivakumar Vaithyanathan: Thumbs up? Sentiment Classification using Machine Learning Techniques CoRR cs.CL/0205070: (2002)
12. Kushal Dave, Steve Lawrence, David M. Pennock: Mining the peanut gallery: opinion extraction and semantic classification of product reviews. WWW 2003: 519-528
13. Christopher Scaffidi, Kevin Bierhoff, Eric Chang, Mikhael Felker, Herman Ng, Chun Jin: Red Opal: product-feature scoring from reviews. ACM Conference on Electronic Commerce 2007: 182-191
14. Minqing Hu, Bing Liu: Mining Opinion Features in Customer Reviews. To appear in AAAI'04, 2004.
15. Jeonghee Yi, Wayne Niblack: Sentiment Mining in WebFountain. ICDE 2005: 1073-1083
16. Roland Siegwart, Illah R. Nourbakhsh, "Introduction to Autonomous Mobile Robots (Intelligent Robotics and Autonomous Agents series) " MIT Press, 2004
17. R. Smith, M. Self, and P. Cheeseman, "A stochastic map for uncertain spatial relationships," in 4th Int. Symp. Robotics Research, O. Faugeras and G. Giralt, Eds., pp. 467–474. The MIT Press, 1988.
18. T. Bar-Shalom and T.E. Fortmann, Tracking and Data Association, Academic Press In., Boston, Mass., 1988.
19. José Neira. SLAM simulation,   http://webdiis.unizar.es/~neira/software/slam/slamsim.htm
20. McGraw, K.L., and B.K. Harbison-Briggs, Knowledge Acquisition, Principles and Guidelines. Upper Saddle River, NJ: Prentice Hall. 1989
21. Dafydd Gibbon, "Computational lexicography," in Lexicon Development for speech and language processing, Frank Van Eynde and Dafydd Gibbon, Eds. Kluwer Academic Publishers, 2000.
22. Likert, Rensis   "A Technique for the Measurement of Attitudes". Archives of Psychology . vol.140, pp. 1–55. 1932
23. Norman, Geoff  "Likert scales, levels of measurement and the "laws" of statistics". Advances in Health Science Education. Vol 15, issue 5,   pp625-632, 2010
24. Pustet, Regina. *Copulas: Universals in the Categorization of the Lexicon*. Oxford studies in typology and linguistic theory. Oxford University Press, 2005
25. Stump, Gregory T. Inflectional morphology: A theory of paradigm structure. Cambridge studies in linguistics. Cambridge: Cambridge University Pres, 2001
26. Doug Hull, Speeding up an enumerated string search in a data mining application. http://blogs.mathworks.com/videos/2010/01/29/advanced-speeding-up-an-enumerated-string-search-in-a-data-mining-application/
27. Swapna Somasundaran and Janyce Wiebe, Recognizing Stances in Ideological On-line Debates In Proceedings of the NAACL HLT 2010 Workshop on Computational Approaches to Analysis and Generation of Emotion in Text, pages 116-124, Los Angeles, CA. Association for Computational Linguistics, 2010.